\documentclass[runningheads]{llncs}

\usepackage{eccv}

\usepackage{eccvabbrv}

\usepackage{graphicx}
\usepackage{booktabs}
\usepackage{epsfig}
\usepackage{amsmath}
\usepackage{amssymb}
\usepackage{wrapfig}
\usepackage{diagbox}
\usepackage{multirow}
\usepackage{bbm}
\usepackage{arydshln}
\usepackage{bbding}

\usepackage{arydshln}

\usepackage{listings}
\usepackage{algorithm}
\usepackage{algorithmic}
\usepackage{setspace}
\usepackage{bm}
\usepackage{tablefootnote}
\usepackage[accsupp]{axessibility}  

\usepackage[breaklinks,colorlinks,citecolor=eccvblue]{hyperref}

\begin{document}

\title{An Efficient and Effective Transformer Decoder-Based Framework for Multi-Task Visual Grounding} 

\titlerunning{EEVG}

\author{Wei Chen\inst{1} \and
Long Chen\inst{2} \and
Yu Wu\inst{1}\thanks{Corresponding author. }}

\authorrunning{Chen et al.}

\institute{$^1$Wuhan University \quad  $^2$The Hong Kong University of Science and Technology
\\
\email{\{weichencs,wuyucs\}@whu.edu.cn, longchen@ust.hk}}

\maketitle

\begin{abstract}

Most advanced visual grounding methods rely on Transformers for visual-linguistic feature fusion. However, these Transformer-based approaches encounter a significant drawback: the computational costs escalate quadratically due to the self-attention mechanism in the Transformer Encoder, particularly when dealing with high-resolution images or long context sentences. This quadratic increase in computational burden restricts the applicability of visual grounding to more intricate scenes, such as conversation-based reasoning segmentation, which involves lengthy language expressions. 
In this paper, we propose an efficient and effective multi-task visual grounding (EEVG) framework based on Transformer Decoder to address this issue, which reduces the cost in both language and visual aspects. In the \textbf{language aspect}, we employ the Transformer Decoder to fuse visual and linguistic features, where linguistic features are input as memory and visual features as queries. This allows fusion to scale \textit{linearly} with language expression length. In the \textbf{visual aspect}, we introduce a parameter-free approach to reduce computation by eliminating background visual tokens based on attention scores. We then design a light mask head to directly predict segmentation masks from the remaining sparse feature maps. Extensive results and ablation studies on benchmarks demonstrate the efficiency and effectiveness of our approach. Code is available in \url{https://github.com/chenwei746/EEVG}. 


\end{abstract}
  
\section{Introduction}
\label{sec:intro}

\begin{figure} 
    \centering  
    \includegraphics[width=1\linewidth]{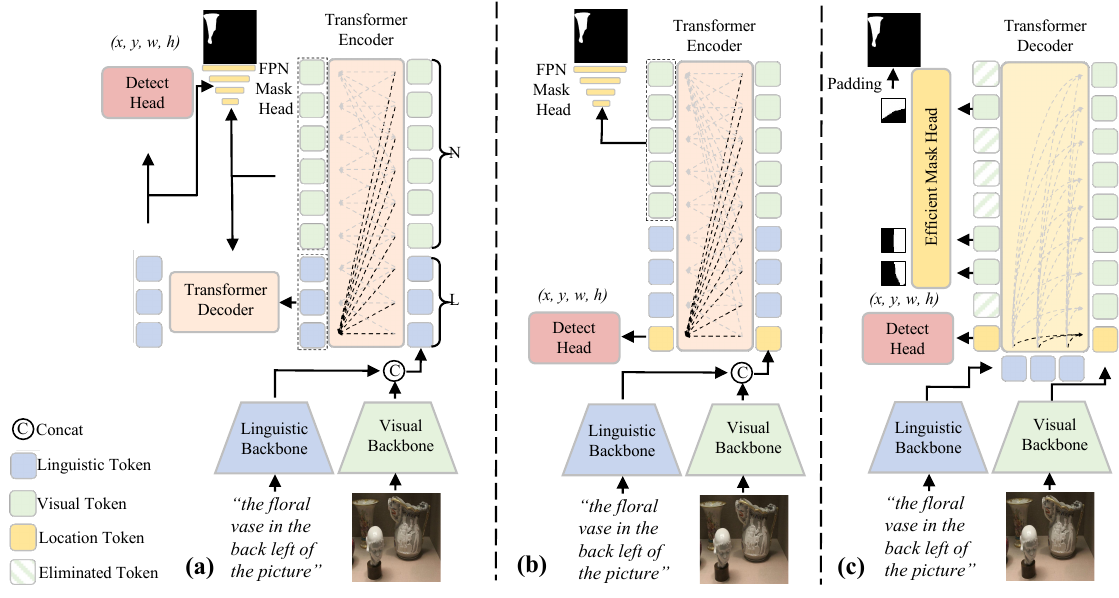}  
    \caption{Comparison of different frameworks: (a) Encoder-Decoder methods, (b) Encoder-only methods, and (c) our Decoder-only framework EEVG. In (a) and (b), Transformer Encoder is utilized, and all visual tokens are employed for mask generation. (c) Our method EEVG leverages the Transformer Decoder to integrate diverse modality information and remove background visual tokens during modalities fusion.}       
    \label{fig:fusion}  
\end{figure}

Visual grounding~\cite{refcoco_and_plus,mao2016generation} is a task of locating visual objects based on language expressions, achieved by aligning visual features and linguistic features. According to the granularity of alignment between visual and linguistic information, it can be categorized into two sub-tasks: referring expression comprehension (REC)~\cite{chen2021ref,deng2021transvg,zhou2021trar,qu2022siri} which facilitates visual-language alignment at the region level, and referring expression segmentation (RES)~\cite{yang2022lavt, wang2022cris,qu2023learning} which grounds language expression at the pixel level. Inspired by joint object detection and segmentation, several works~\cite{luo2020multi,li2021referring,zhu2022seqtr, liu2023polyformer,cheng2023pvd,su2023vglaw} propose a multi-task collaborative learning framework to unify REC and RES, \ie, multi-task visual grounding (MTVG). They demonstrate that REC aids RES in locating referents more accurately, while RES helps REC achieve better vision-language alignment. As a result, MTVG has become a prevailing way of visual grounding.

Current mainstream MTVG models consist of a visual encoder, a linguistic encoder, a cross-modal feature fusion module, and two task heads (\textit{i.e.}, a detection head and a segmentation mask head). 
Recent state-of-the-art visual grounding works~\cite{li2021referring,deng2021transvg,kamath2021mdetr,kim2022restr,liu2023polyformer} resort to Encoder\footnote{We use \emph{Encoder} and \emph{Decoder} to refer to Transformer encoder and decoder, respectively.} for cross-modal vision-language fusion. As shown in Fig.~\ref{fig:fusion}, these existing methods can be categorized into two categories: (a) Encoder-Decoder~\cite{li2021referring,kamath2021mdetr,qu2022siri,liu2023polyformer}, where visual-linguistic features are fused by Encoder and target object features are output by Decoder, and (b) Encoder-only~\cite{deng2021transvg,kim2022restr}, where the target object is directly predicted after vision-language feature fusion in Encoder.

However, these methods encounter two efficiency problems: 1) quadratic-increased cost in language length and 2) redundant visual token computation. \textit{Firstly}, traditional methods concatenate visual and linguistic tokens together and input them into Encoder for self-attention, leading to a time complexity of $\mathcal{O}((N+L)^2)$. Thus the computation cost significantly increases as language expressions and context become longer and more complex in the era of Large Language Models (LLMs). This hinders the application of visual grounding to more complex scenes, such as conversation-based reasoning segmentation~\cite{lai2023lisa}, which involves long language contexts. 
\textit{Secondly}, different from general segmentation or detection tasks, visual grounding usually only aims at locating one referred object. Most visual pixels in the image are not in the region of interest and thus lead to redundant and unnecessary computations, and may distract/mislead the model's attention from the real target. To address these issues, we propose an efficient and effective multi-task visual grounding (EEVG) framework. 

To alleviate the cost in the \textbf{language aspect} and to deal with longer complex language expressions such as long contextual dialog, we only use Decoder for the visual and language reasoning process. As depicted in Fig.~\ref{fig:fusion}~(c), we regard linguistic features as memory and visual features as queries in Decoder. 
This allows for the fusion of visual and linguistic modalities in the cross-attention module, resulting in a \textit{linear} increase in computational cost with respect to the length of the language expressions. 
To the best of our knowledge, our
method is the first Transformer-based framework with \textit{no Encoder} for cross-modal fusion.

To mitigate the cost in the \textbf{visual aspect} and further improve efficiency and efficacy, we introduce a parameter-free strategy to eliminate redundant and distracting image tokens for the visual grounding task.
The core idea is dynamically eliminating visual tokens with low attention scores, making the visual feature map to be sparse, and concentrating more on the referred target to remove distracting noise.
After that, instead of utilizing the traditional feature pyramid network (FPN)~\cite{kirillov2019panoptic} that is widely used in previous MTVG works, we devise a very light-weight and efficient mask head to directly project the remaining sparse tokens into region masks. 
Previous works'~\cite{huang2020referring,feng2021encoder,jing2021locate,li2021referring,yang2022lavt,kim2022restr,su2023vglaw} FPN module constitutes 42.8\% of the Decoder's parameters (8.1M \textit{versus} 18.9M), acting like an independent network to predict mask.
Differently, we use a light-weight two-layer MLP (0.79M) to directly transfer 1-D feature channels of Decoder tokens to the 2-D spatial segmentation mask prediction of the corresponding patch.
In this way, we migrate the segmentation prediction workload from the add-on head to the main Decoder.
This is consistent with the detection head of REC which also transfers the detection workload to Decoder via a light-weight MLP.

By doing so, Decoder gains a better understanding of the multi-tasks and their mutual improvement, as the location and pixel information are directly embedded in the Decoder token feature.
Experiments validate this by the fact that our mask head improves the detection performance on REC by about 2.0\%, even though the detection prediction does not go through the mask head.
We conduct extensive experiments on several challenging benchmarks including RefCOCO~\cite{refcoco_and_plus}, RefCOCO+~\cite{refcoco_and_plus}, and RefCOCOg~\cite{mao2016generation}. Our EEVG is faster than state-of-the art method PolyFormer~\cite{liu2023polyformer} by 28.19\%. Benefiting from light-weight mask head and elimination of disturbing tokens, EEVG also shows enhanced performance. Particularly in the RefCOCOg dataset, which encompasses longer complex language expressions, our method exhibits a notable increase of 3.93\% on the RES. 

In summary, our contributions are three-fold: 
\begin{itemize}
    \item We propose a Decoder-only framework for MTVG, which reduces computation cost from quadratic to linear increase with regards to language length.
  
    \item We propose a dynamic eliminating strategy to reduce redundant and distracting visual tokens, together with a lightweight mask head to directly project the remaining sparse tokens to masks.
    
    \item Comprehensive results show that EEVG surpasses state-of-the-art approaches in both speed and performance.
\end{itemize}

\section{Related Work}







\noindent\textbf{Referring Expression Comprehension (REC). }
REC can be categorized into one-stage and two-stage approaches. Two-stage~\cite{hu2017modeling, zhang2018grounding, hong2019learning} methods rely on ranking region proposal scores based on language expressions as a crucial component. However, their performance is limited to the pre-trained object detector. 
While one-stage methods~\cite{yang2019fast,yang2020improving,liao2020real, kamath2021mdetr} focus on directly predicting the target bounding box guided by language expressions. Yang \etal~\cite{yang2020improving} propose a recursive sub-query construction framework to reason between image and query for multiple rounds. To further align modalities, RCCF~\cite{liao2020real} maps the language domain to the visual domain and performs correlation filtering on the image feature map. Inspired by DETR's~\cite{carion2020end} success in object detection, MDETR~\cite{kamath2021mdetr} extends it to multi-modal understanding for REC.

\noindent\textbf{Referring Expression Segmentation (RES). }
As RES needs to predict pixel-level results, it heavily relies on accurate vision-language feature extraction and alignment. Previous studies~\cite{huang2020referring,feng2021encoder,jing2021locate,yang2022lavt} have explored various approaches for cross-modal interaction. EFN~\cite{feng2021encoder} utilizes a co-attention mechanism to promote the consistency of the cross-modal information representation in the semantic space. On the other hand, LTS~\cite{jing2021locate} leverages visual-textual features to accurately localize the referenced object by incorporating position priors before facilitating segmentation. Recent work LAVT~\cite{yang2022lavt} aligns visual and linguistic representations within the visual backbone using a pixel-word attention module. Current approaches~\cite{huang2020referring,feng2021encoder,jing2021locate,li2021referring,yang2022lavt,kim2022restr,su2023vglaw} typically employ an FPN-like architecture to generate binary masks from fused visual features. In contrast, our proposed method introduces a lighter mask head based on MLP.

\noindent\textbf{Multi-task Visual Grounding (MTVG). }
To promote consistency between REC and RES, it is natural to integrate them using a shared backbone. In a pioneering effort, MCN~\cite{luo2020multi} proposes a novel multi-task collaborative network that enables joint learning of REC and RES. With the widespread adoption of Transformer~\cite{vaswani2017attention}, follow-up works~\cite{li2021referring,kim2022restr,su2023vglaw} have employed Transformer as a unified backbone, employing different task heads for REC and RES. Alternatively, SeqTR~\cite{zhu2022seqtr} approaches the problem differently by treating MTVG as a sequence prediction task, representing bounding boxes and masks as discrete coordinate tokens. Moreover, Polyformer~\cite{liu2023polyformer} leverages precise floating-point coordinates and multi-polygon generation to achieve finer segmentation. Tasks similar to MTVG, Open-Vocabulary Object Detection/Segmentation~\cite{gu2021open, xu2023side} need to identify all objects across all categories using the vocabularies.

\begin{figure}[!htbp]
\centering
\begin{minipage}[t]{0.48\textwidth}
    \centering
    \vfill
    \includegraphics[width=1\linewidth]{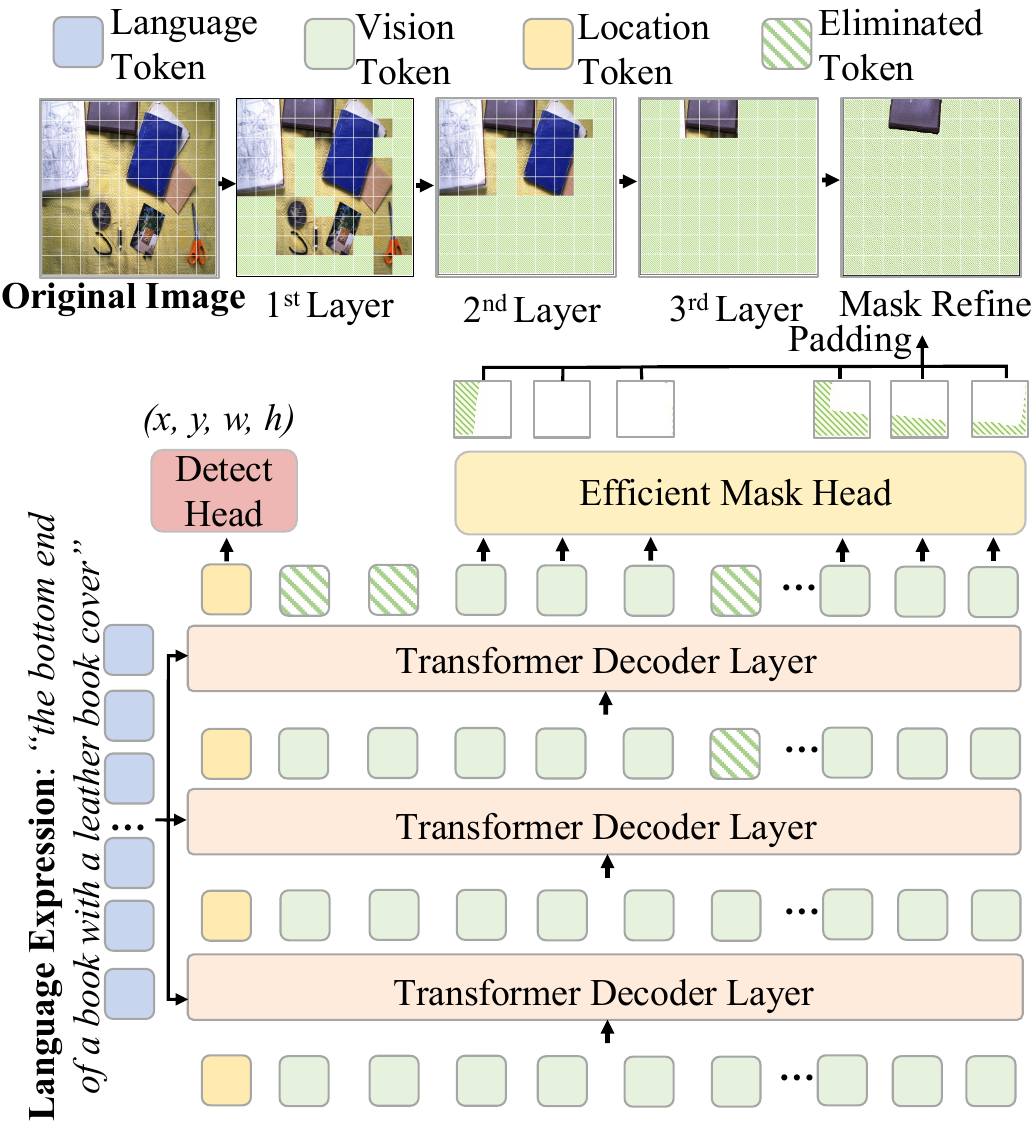}
    \caption{Overview of our method. The language tokens and visual tokens are extracted by a linguistic backbone and a visual backbone which are not shown in the figure. 
    }
    \label{fig:method}
\end{minipage}
\hspace{2mm}
\begin{minipage}[t]{0.48\textwidth}
    \vfill
    \includegraphics[width=1\linewidth]{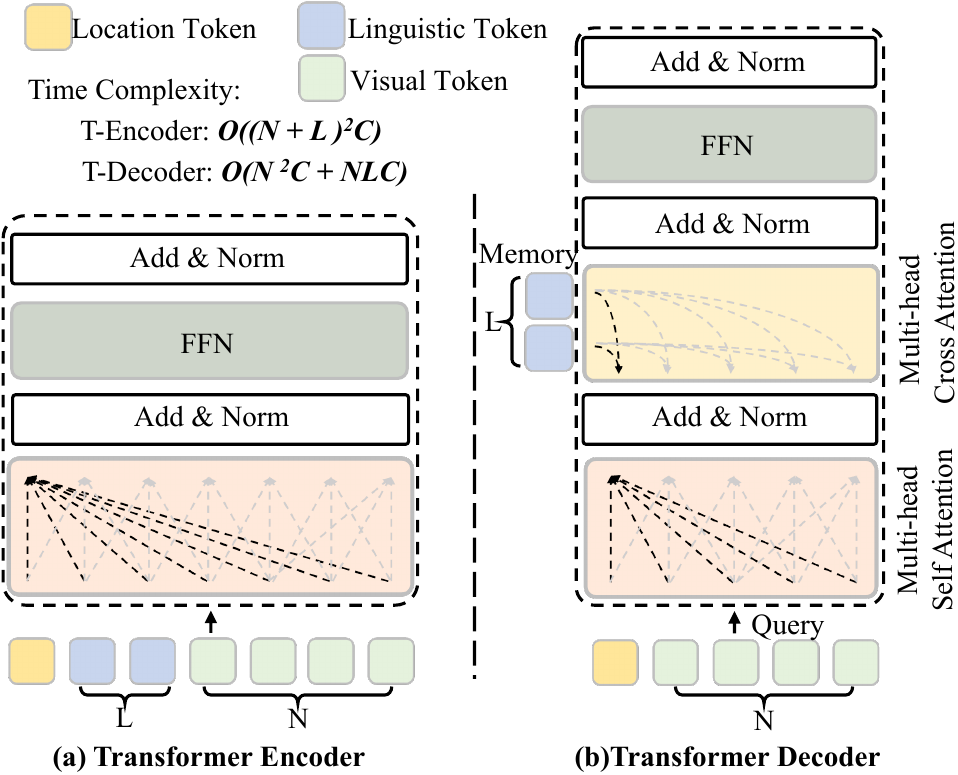}
    \caption{Time complexity comparison between Encoder and Decoder. There is only one input for Encoder while Decoder has two inputs: query and memory. $N$ denotes the number of visual tokens, $L$ means the number of linguistic tokens, $C$ is the dimension of tokens, and ``Add \& Norm'' refers to residual connection and normalization. }
    \label{fig:encoder_decoder}
\end{minipage}
\end{figure}

\noindent\textbf{Transformer for Vision-Language Tasks. }
Transformer model~\cite{vaswani2017attention}, initially proposed for natural language processing tasks, has demonstrated its effectiveness in the computer vision domain as well, as evidenced by the success of Vision Transformer (ViT)~\cite{dosovitskiy2020image,wang2023crossformer++}. Leveraging the Transformer's exceptional performance in both vision and natural language, researchers have extensively explored its potential as a unified model for vision-language tasks~\cite{lu2019vilbert,chen2020uniter,wang2022ofa}. 
Recently, there have been a growing number of methods~\cite{deng2021transvg,li2021referring,zhu2022seqtr,liu2023polyformer,su2023vglaw} based on the Transformer architecture in visual grounding. 
TransVG~\cite{deng2021transvg}, which employs Transformer Encoder for cross-modal fusion, introduces the pioneering transformer-based framework for visual grounding.

\section{Method}

In this section, we first formulate the multi-task visual grounding (MTVG) task and review the prevalent Encoder-based framework in Sec.~\ref{sub_sec:preliminary}. Then, we elaborate on our new Decoder-based framework as shown in Fig.~\ref{fig:method}, which utilizes Decoder for vision-language fusion (Sec.~\ref{sub_sec:Decoder}), and includes a parameter-free strategy to eliminate visual tokens (Sec.~\ref{sub_sec:token_elimination}) and an efficient mask head (Sec.~\ref{sub_sec:mask_head}).

\subsection{Preliminary}\label{sub_sec:preliminary}

\textbf{Formultation.} Given an image $\mathcal{I} \in \mathbb{R}^{H \times W \times 3}$ and a text query $\mathcal{T} \in \mathbb{R}^{L}$, MTVG needs to predict a bounding box $B$ and a binary mask $M$ simultaneously, which corresponding to the referent. 
The current prevailing MTVG framework, typically, first utilizes a visual backbone (\eg, ViT~\cite{dosovitskiy2020image}) and a linguistic backbone (\eg, BERT~\cite{devlin2018bert}) to extract visual features $\bm{F}_v \in \mathbb{R}^{N \times C_v}$ and linguistic features $\bm{F}_l \in \mathbb{R}^{L \times C_l}$. After fusing them in the cross-modal interaction module, two task heads are used to predict results.

\textbf{Encoder-based Framework.} Existing works~\cite{deng2021transvg,kim2022restr,liu2023polyformer} directly adopt Encoder as the cross-modal interaction module. As shown in Fig.~\ref{fig:encoder_decoder}~(a), after linearly projecting $\bm{F}_v$ and $\bm{F}_l$ into the same dimension $C$ ($\bm{\widetilde{F}}_v$ and $\bm{\widetilde{F}}_l$), $\bm{\widetilde{F}}_v$ and $\bm{\widetilde{F}}_l$ are concatenated with a learnable location token $\bm{\widetilde{F}}_{loc} \in \mathbb{R}^{1 \times C}$ and fed into Encoder:
\begin{equation}
    [\hat{\bm{F}}_{loc}, \hat{\bm{F}}_{l}, \hat{\bm{F}}_{v}] = \texttt{Encoder}([\bm{\widetilde{F}}_{loc}; \bm{\widetilde{F}}_{l}; \bm{\widetilde{F}}_{v}]),
\end{equation}
where $[~\cdot;~\cdot;~\cdot]$ denotes the concatenation operation. 

After getting fused features, for the detection task, a two-layer MLP is used to project $\hat{\bm{F}}_{loc}$ into four dimensions (\ie, $(x, y, w, h)$). For the segmentation results, $\hat{\bm{F}}_{v}$ is reshaped from sequence to square (\ie, $\mathbb{R}^{N \times C} \rightarrow \mathbb{R}^{\frac{H}{P} \times \frac{W}{P} \times C}$ where $P$ is the patch size) and uses FPN-like~\cite{kirillov2019panoptic} architecture built on convolution layers to generate masks. This architecture typically needs the entire visual features because convolution layers can only work in a square feature map which contains redundant costs in the visual features corresponding to the background. Therefore, we utilize our elimination strategy and efficient head to alleviate this. 

\textbf{Time Complexity Analysis. } The computational cost of aligning different modalities in the Encoder-based methods primarily lies in the multi-head self-attention (MSA) which can be formulated as:
\begin{equation}
    \texttt{MSA}(\bm{Q}, \bm{K}, \bm{V}) = \texttt{softmax}(\frac{\bm{QK}^T}{\sqrt{C}})\bm{V}, \label{eq:attn}
\end{equation}
where $\bm{Q}$, $\bm{K}$, and $\bm{V}$ represent query, key, and value, respectively. They are obtained through linear projections of the input features. Specifically, $\bm{Q}$, $\bm{K}$, and $\bm{V} \in \mathbb{R}^{(N+L+1) \times C}$. According to Eq.~\eqref{eq:attn}, its time complexity can be calculated as $\mathcal{O}((N+L)^{2}C)$. In our method, we aim to alleviate the quadratic increase in burden by reducing it to a linear one with respect to $L$, as well as minimizing the number of $N$.

\subsection{Transformer Decoder for Modalities Fusion} \label{sub_sec:Decoder}

\textbf{Module Details. } First, we separately project visual feature $\bm{F}_v$ and linguistic feature $\bm{F}_l$ into the same channel dimension $C$ ($\bm{\widetilde{F}}_v$ and $\bm{\widetilde{F}}_l$). 
Then we adopt Transformer Decoder~\cite{vaswani2017attention} for vision-language fusion, as depicted in Fig.~\ref{fig:encoder_decoder}~(b), where each decoder layer consists of an MSA layer, a multi-head cross-attention (MCA) layer, and a feed-forward network. MCA has a similar architecture to MSA, but the difference is that MCA has two inputs: the first one is input as Q and another one is input as K and V in Eq.~\eqref{eq:attn}.  We concatenate $\bm{\widetilde{F}}_v$ with a learnable location token $\bm{\widetilde{F}}_{loc}$ and input them into the MSA layer: 
\begin{equation}
\begin{aligned}
    \relax[\bm{F}_{loc}^{\prime}, \bm{F}_{v}^{\prime}] &= \texttt{LN} ( \texttt{MSA} ([\bm{\widetilde{F}}_{loc}; \bm{\widetilde{F}}_{v}]) + [\bm{\widetilde{F}}_{loc}; \bm{\widetilde{F}}_{v}] ),
\end{aligned}
\end{equation}
where $\texttt{LN} ( \cdot )$ refers to layer normalization. After that, we input $[\bm{F}_{loc}^{\prime}; \bm{F}_{v}^{\prime}]$ as the query of MCA and input linguistic feature $\bm{\widetilde{F}}_{l}$ as the key and value of MCA: 
\begin{equation}
    [\bm{F}_{loc}^{*}, \bm{F}_{v}^{*}] = \texttt{LN} ( \texttt{MCA}([\bm{F}_{loc}^{\prime}; \bm{F}_{v}^{\prime}], \bm{\widetilde{F}}_{l}) + [\bm{F}_{loc}^{\prime}; \bm{F}_{v}^{\prime}] ),
\end{equation}

Finally, $[\bm{F}_{loc}^{*}, \bm{F}_{v}^{*}]$ is passed into the feed-forward network:
\begin{equation}
    [\hat{\bm{F}}_{loc}, \hat{\bm{F}}_{v}] = \texttt{LN} (\texttt{FFN}([\bm{F}_{loc}^{*}; \bm{F}_{v}^{*}]) + [\bm{F}_{loc}^{*}; \bm{F}_{v}^{*}]),
\end{equation}
where $\texttt{FFN} (\cdot)$ means the feed-forward network, which is a two-layer MLP.

\textbf{Time Complexity Analysis.} The computational cost mainly lies in the MSA and MCA, with time complexities of $\mathcal{O}(N^{2}C)$ and $\mathcal{O}(NLC)$, respectively. Thus, the overall time complexity is $\mathcal{O}(N^{2}C + NLC)$, which increases linearly with respect to $L$.

\subsection{Parameter-free Token Elimination Strategy}\label{sub_sec:token_elimination}

\setlength{\intextsep}{0pt}
\begin{wrapfigure}[18]{r}{6cm}
    \centering
    \includegraphics[width=1\linewidth]{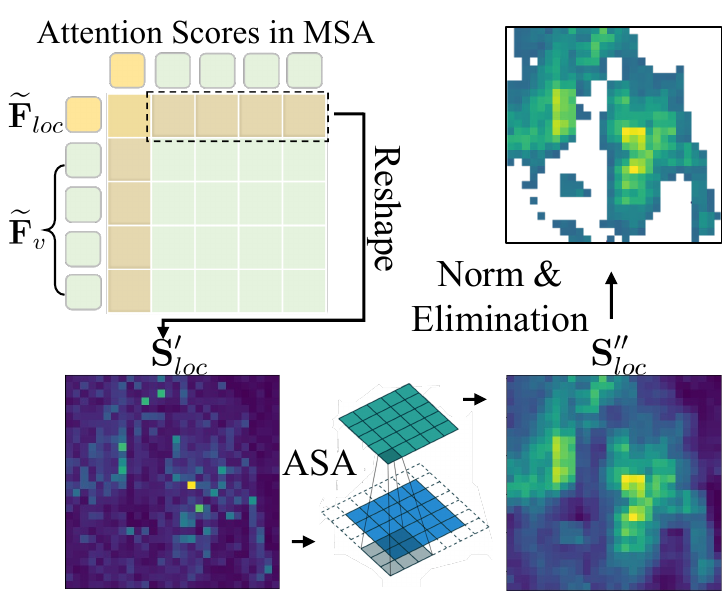}
    \caption{We conduct the visual tokens elimination process in each Decoder layer. ``ASA'' denotes adaptive spatial attention and ``Norm \& Elimination'' means normalization and eliminating visual tokens according to Eq.~\eqref{eq:normalization}.     }
    \label{fig:token_eli}
\end{wrapfigure}
Visual grounding usually aims at locating one referent and most referents only occupy a small percentage of the visual tokens where most visual tokens are not in the region of interest. Therefore, there exist redundant costs in background visual tokens. As a result, we attempt to address this issue by eliminating background visual tokens.
During our analysis, we discovered that the attention scores between the location token and the visual tokens are notably higher for those corresponding to the target object. This 
finding led us to the conclusion that we can effectively eliminate visual tokens with low attention scores which are depicted in Fig.~\ref{fig:token_eli}.

\noindent\textbf{Dynamic Elimination. }
There are some related works~\cite{fayyaz2022adaptive,bolya2022token} eliminating visual tokens in image classification. However, these methods eliminate a fixed number in each model layer and typically retain only the essential object features. These approaches are suitable for coarse granularity tasks like image classification. In contrast, the task of RES requires the preservation of the entire object and deals with objects of varying sizes, each demanding a different number of tokens to be eliminated. To tackle this challenge, we propose a dynamic elimination strategy. First, attention scores $\bm{S}_{loc} \in \mathbb{R}^{1 \times N}$ between the location token and visual tokens are calculated as: 
\begin{equation}
\begin{aligned}
    \bm{Q}_{loc} &= \bm{\widetilde{F}}_{loc}\bm{W}_{\bm{Q}}, \bm{K}_{v} = \bm{\widetilde{F}}_{v}\bm{W}_{\bm{K}}, \\
    \bm{S}_{loc} &= \texttt{softmax} (\frac{\bm{Q}_{loc}\bm{K}_{v}^T}{\sqrt{C}}),
\end{aligned}
\end{equation}
where $\bm{W}_{\bm{Q}} \in \mathbb{R}^{C \times C}$ and $\bm{W}_{\bm{K}} \in \mathbb{R}^{C \times C}$ are parameter matrices in MSA. Then we normalize the attention scores $\bm{S}_{loc}$ and remove those visual tokens whose attention scores are smaller than $\alpha$:
\begin{equation}
\begin{aligned}
    \bm{\widetilde{S}}_{loc} &= \frac{\bm{S}_{loc} - \min \bm{S}_{loc}^{i}}{\max \bm{S}_{loc}^{i} - \min \bm{S}_{loc}^{i}}, 0 \leq i < N \\
    \overline{\bm{F}}_{v} &= \{ \hat{\bm{F}}^{i}_{v} \mid \bm{\widetilde{S}}_{loc}^{i} \geq \alpha,~0 \leq i < N \},
\end{aligned}
\label{eq:normalization}
\end{equation}
where $\hat{\bm{F}}^{i}_{v}$ is the $i$-th visual token, $\overline{\bm{F}}_{v} \in \mathbb{R}^{N^{\prime} \times C}$ denotes the remained tokens, and $N^{\prime}$ means the number of remained visual tokens. Our strategy offers two advantages over prior works~\cite{fayyaz2022adaptive,bolya2022token}: 1) it dynamically eliminates a different number of tokens for different objects in different sizes, and 2) the number of eliminated tokens gradually increases as the loss converges, preventing error elimination.

\begin{figure}[t]
    \centering
    \includegraphics[width=1\linewidth]{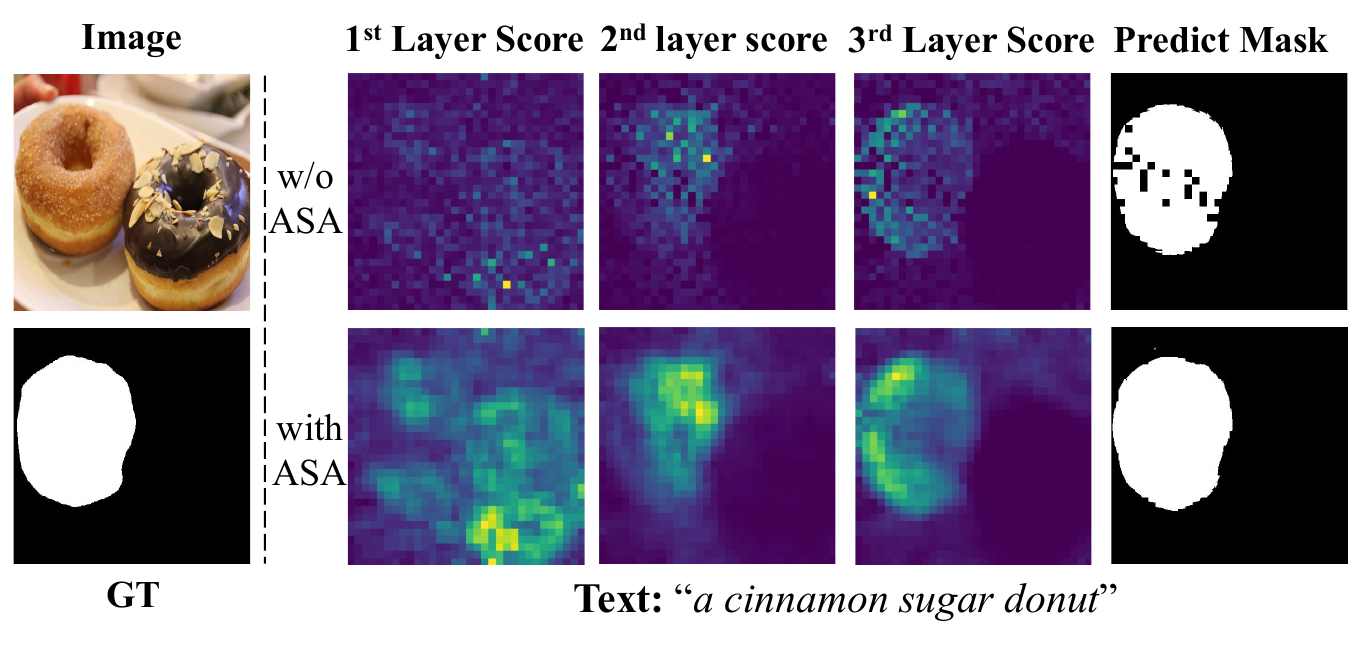}
    \caption{Visualization of attention scores $S_{loc}$ of location token and visual tokens. GT refers to ground truth, ``w/o ASA'' denotes without using adaptive spatial attention, and ``$1^{st} \text{Layer Score}$'' means the first Decoder layer attention scores between location token and visual tokens. }
    \label{fig:attention_score}
\end{figure}

\noindent\textbf{Adaptive Spatial Attention. }
As shown in the predicted mask in Fig.~\ref{fig:attention_score}, 
we found that some visual tokens corresponding to target objects are eliminated incorrectly.
Therefore, we propose the attention weight average strategy to make the attention weights spatially aware to maintain the original shape of the target object. We reshape $\bm{S}_{loc}$ into $\bm{S}_{loc}^{\prime} \in \mathbb{R}^{\frac{H}{P} \times \frac{W}{P}}$, then calculate as follows:
\begin{equation}
    \bm{S}_{loc}^{\prime\prime}[i, j] = \frac{\displaystyle\sum_{u=-k}^{k}\displaystyle\sum_{v=-k}^{k} \bm{S}_{loc}^{\prime}[i+u, j+v]}{(2k+1)^2},
\end{equation}
where $0 \leq i < \frac{H}{P}$ and $0 \leq j < \frac{W}{P}$. 
After that, each attention score is averaged with the surrounding attention scores, then we reshape $\bm{S}_{loc}^{\prime\prime}$ back to $\bm{S}_{loc} \in \mathbb{R}^{1 \times N}$ and following Eq.~\eqref{eq:normalization} to eliminate visual tokens. So we can alleviate the problem of incorrectly eliminating tokens in the referent.

\subsection{Efficient Mask Head} \label{sub_sec:mask_head}
\begin{figure}[t]
    \centering
    \includegraphics[width=1\linewidth]{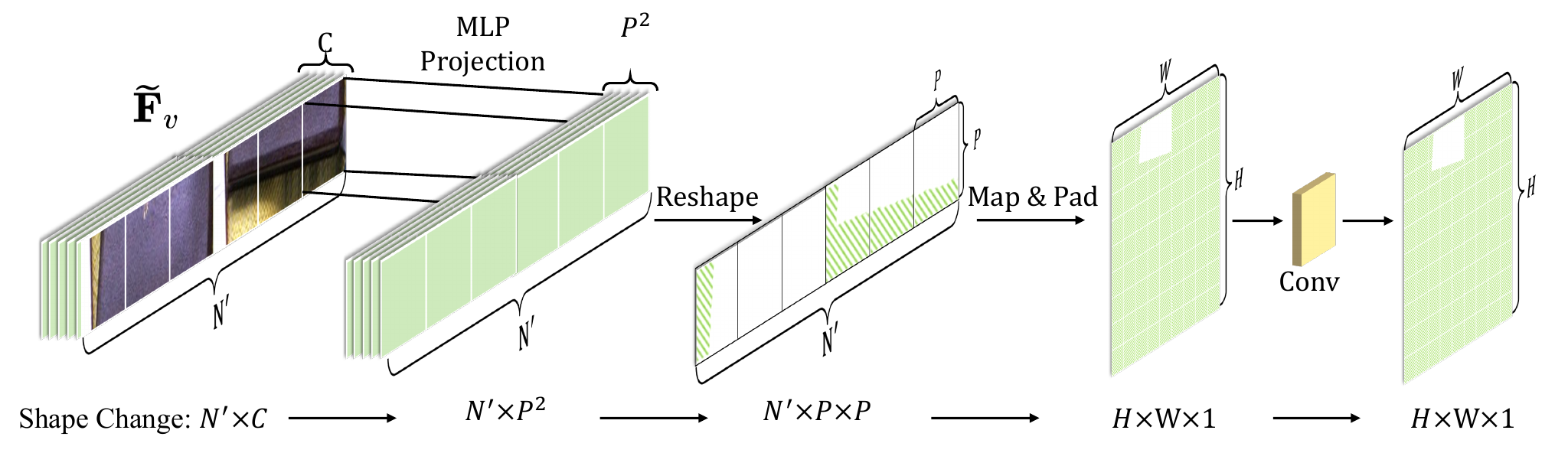}
    \caption{Our mask head utilizes MLP to project tokens from channel to spatial region masks and adopts a 1-channel convolution layer to establish spatial relationships among pixels. ``Map \& Pad'' refers to mapping each patch to the original position in the image and padding the eliminated position with 0. 
    }
    \label{fig:token_eli_and_mask_head}
\end{figure}

After eliminating tokens, we get some sparse visual tokens. To further reduce the redundant cost, we propose a lightweight and efficient mask head to generate masks from sparse visual tokens instead of using the FPN-like mask head which needs to pad the eliminated tokens. The procedure of our mask head is shown in Fig.~\ref{fig:token_eli_and_mask_head}. We denote the rest of indexes $\mathbb{I}$ as follows:
\begin{equation}
    \mathbb{I} = \{ i \mid \bm{\widetilde{S}}_{loc}^{i} \geq \alpha,~0 \leq i < N \},
\end{equation}
We utilize MLP to transfer the remaining visual tokens from 1-D feature channels into 2-D spatial binary masks ($C \rightarrow P^2$ dimension) and pad the eliminated tokens with 0:
\begin{equation}
    \mathcal{M}^{i} = 
    \begin{cases}
    MLP( \overline{\bm{F}}_{v}^{f(i)} ), &\text{if $i \in \mathbb{I}$}\\
    0^{1 \times P^{2}}, &\text{if $i \notin \mathbb{I}$}
    \end{cases}
\end{equation}
where $0 \leq i < N$ and $f(i)$ means the index in remained tokens corresponding to the $i$-th original token. We reshape and permute $\mathcal{M} \in \mathbb{R}^{N \times P^2}$ into $\mathcal{M}^{\prime} \in \mathbb{R}^{H \times W}$ and since the pixels projected by MLP are not spatially related to each other, we use the local context processing method to make neighboring pixels relate to each other. In the actual implementation, we use a 5*5 convolutional kernel $Conv$, and mask $M$ is generated as follows:
\begin{equation}
    M = \texttt{sigmoid}(\texttt{Conv}(\mathcal{M}^{\prime})),
\end{equation}

\subsection{Multi-task Training}
Our method is an end-to-end framework that unifies REC and RES, incorporating two distinct types of loss: detection loss and segmentation loss.

\textbf{Detection Loss.} For REC task, we denote the predicted bounding box $B = (\hat{x}, \hat{y}, \hat{w}, \hat{h})$ and the ground truth $B_{gt} = (x, y, w, h)$, and the detection loss is define as follows:
\begin{equation}
    \mathcal{L}_{det} = \mathcal{L}_{smooth-L1}(B, B_{gt}) + \mathcal{L}_{giou}(B, B_{gt}),
\end{equation}
where $\mathcal{L}_{smooth-L1}(\cdot,\cdot)$ and $\mathcal{L}_{giou}(\cdot,\cdot)$ are the smooth L1 loss and GIoU loss~\cite{rezatofighi2019generalized}. 

\textbf{Segmentation Loss. }For RES task, the segmentation loss is calculated using the predicted segmentation mask denoted as $M \in \mathbb{R}^{H \times W}$, and the ground truth denoted as $M_{gt} \in \mathbb{R}^{H \times W}$, according to the following formula:
\begin{equation}
    \mathcal{L}_{seg} = \mathcal{L}_{focal}(M, M_{gt}) + \mathcal{L}_{dice}(M, M_{gt}),
\end{equation}
where $\mathcal{L}_{focal}(\cdot,\cdot)$ and $\mathcal{L}_{dice}(\cdot,\cdot)$ represent focal loss~\cite{lin2017focal} and dice loss~\cite{milletari2016v}. Finally, the joint training loss function is defined as follows:
\begin{equation}
    \mathcal{L} = \lambda_{det}\mathcal{L}_{det} + \lambda_{seg}\mathcal{L}_{seg}.
\end{equation}
where $\lambda_{det}$ and $\lambda_{seg}$ represent the weight coefficients for the detection loss and segmentation loss, respectively.

    \begin{table*}[t]
\centering
\scriptsize
\caption{Comparison with state-of-the-art methods on RefCOCO~\cite{refcoco_and_plus}, RefCOCO+~\cite{refcoco_and_plus}, and RefCOCOg~\cite{nagaraja2016modeling} for \textbf{REC} task. \textbf{Bold} denotes the best performance. Swin-B and ViT-B are abbreviations for Swin-Transformer Base and ViT Base.}

\begin{tabular}{c|c|c|ccc|ccc|cc}
\noalign{\hrule height 1.5pt}
\multicolumn{1}{c|}{\multirow{2}{*}{Method}} & \multicolumn{1}{c|}{\multirow{2}{*}{Backbone}} & Multi- &\multicolumn{3}{c|}{RefCOCO} & \multicolumn{3}{c|}{RefCOCO+} & \multicolumn{2}{c}{RefCOCOg} \\
\cline{4-11}
& & task &val & test A & test B & val & test A & test B & val(U) & test(U) \\
\hline
MAttNet~\cite{yu2018mattnet} & MRCNN-Res101&\XSolidBrush  & 76.65 & 81.14 & 69.99 & 65.33 & 71.62 & 56.02 & 66.58 & 67.27  \\
NMTree~\cite{liu2019learning} & MRCNN-Res101&\XSolidBrush & 76.41 & 81.21 & 70.09 & 66.46 & 72.02 & 57.52 & 65.87 & 66.44  \\
LBYL~\cite{huang2021look} & DarkNet53&\XSolidBrush & 79.67 & 82.91 & 74.15 & 68.64 & 73.38 & 59.49 & - & -  \\
MCN~\cite{luo2020multi} & DarkNet53&\Checkmark & 80.08 & 82.29 & 74.98 & 67.16 & 72.86 & 57.31 & 66.46 & 66.01\\
TransVG~\cite{deng2021transvg} & ResNet101&\XSolidBrush & 81.02 & 82.72 & 78.35 & 64.82 & 70.70 & 56.94 & 68.67 & 67.73 \\
TRAR~\cite{zhou2021trar} & DarkNet53&\XSolidBrush & - & 81.40 &  {78.60} & - & 69.10 & 56.10 & 68.90 & 68.30 \\
SeqTR~\cite{zhu2022seqtr} & DarkNet53&\Checkmark & 81.23 & 85.00 & 76.08 & 68.82 & 75.37 & 58.78 &  {71.35} &  {71.58} \\
PVD~\cite{cheng2023pvd} & DarkNet53&\Checkmark &  {82.51} &  {86.19} & 76.81 &  {69.48} &  {76.83} &  {59.68} & 68.40 & 69.57  \\
PVD~\cite{cheng2023pvd} & Swin-B&\Checkmark & 84.52 & 87.64 & 79.63 & 73.89 & 78.41 & 64.25 & 73.81 & 74.13  \\
VG-LAW~\cite{su2023vglaw} & ViT-B&\Checkmark & 86.62 & 89.32 & 83.16 & 76.37 & 81.04 & 67.50 & 76.90 & 76.96  \\
\hline

EEVG~\textbf{(Ours) }  & MRCNN-Res101 &\Checkmark & 82.19 & 85.34 & 77.18 & 71.35 & 76.76 & 60.73 & 70.18 & 71.28 \\

EEVG~\textbf{(Ours) }  & DarkNet53 &\Checkmark & 81.82 & 86.02 & 74.67 & 69.72 & 76.26 & 57.95 & 71.38 & 70.93 \\

EEVG~\textbf{(Ours) }  & Swin-B&\Checkmark & 86.79 & 89.52 & {83.12} & 77.52 & \textbf{83.05} & 66.93 & 78.15 & 78.11 \\

EEVG~\textbf{(Ours) }  & ViT-B&\Checkmark & \textbf{88.08} & \textbf{90.33} & \textbf{85.50} & \textbf{77.97} & {82.44} & \textbf{69.15} & \textbf{79.60} & \textbf{80.24} \\



\noalign{\hrule height 1.5pt}
\end{tabular}

\label{tab:main_rec_det}
\end{table*}

    \begin{table*}[t]
\centering
\scriptsize

\caption{Comparison with state-of-the-art methods on RefCOCO~\cite{refcoco_and_plus}, RefCOCO+~\cite{refcoco_and_plus}, and RefCOCOg~\cite{nagaraja2016modeling} for \textbf{RES} task. \textbf{Bold} denotes the best performance. Swin-B and ViT-B are abbreviations for Swin-Transformer Base and ViT Base.
}

\begin{tabular}{c|c|c|ccc|ccc|cc}
\noalign{\hrule height 1.5pt}
\multicolumn{1}{c|}{\multirow{2}{*}{Method}} & \multicolumn{1}{c|}{\multirow{2}{*}{Backbone}} & Multi- & \multicolumn{3}{c|}{RefCOCO} & \multicolumn{3}{c|}{RefCOCO+} & \multicolumn{2}{c}{RefCOCOg} \\
\cline{4-11}

 & & task & val & test A & test B & val & test A & test B & val(U) & test(U) \\

\hline
MAttNet~\cite{yu2018mattnet} & MRCNN-Res101&\XSolidBrush & 56.51 & 62.37 & 51.70 & 46.67 & 52.39 & 40.08 & 47.64 & 48.61\\
NMTree~\cite{liu2019learning} & MRCNN-Res101&\XSolidBrush &  56.59 & 63.02 & 52.06 & 47.40 & 53.01 & 41.56 & 46.59 & 47.88 \\
MCN~\cite{luo2020multi} & DarkNet53&\Checkmark & 62.44 & 64.20 & 59.71 & 50.62 & 54.99 & 44.69 & 49.22 & 49.40  \\
CRIS~\cite{wang2022cris} & CLIP-ResNet50&\XSolidBrush & 69.52 & 72.72 & 64.70 & 61.39 & 67.10 & 52.48 & 59.87 & 60.36 \\ 
SeqTR~\cite{zhu2022seqtr} & DarkNet53&\Checkmark & 67.26 & 69.79 & 64.12 & 54.14 & 58.93 & 48.19 & 55.67 & 55.64 \\
PVD~\cite{cheng2023pvd} & DarkNet53&\Checkmark & 68.87 & 70.53 & 65.83 & 54.98 & 60.12 & 50.23 & 57.81 & 57.17 \\
LAVT~\cite{yang2022lavt} & Swin-B&\XSolidBrush & 74.46 & 76.89 & 70.94 & 65.81 & 70.97 & 59.23 & 63.34 & 63.62  \\
PVD~\cite{cheng2023pvd} & Swin-B&\Checkmark & 74.82 & 77.11 &  69.52 & 63.38 & 68.60 & 56.92 & 63.13 & 63.62  \\
VG-LAW~\cite{su2023vglaw} & ViT-B&\Checkmark & 75.62 & 77.51 & 72.89 & 66.63 & 70.38 & 59.89 & 65.53 & 66.08 \\
\hline

EEVG~\textbf{(Ours) } & MRCNN-Res101&\Checkmark & 71.28 & 73.87 & 67.49 & 61.96 & 66.25 & 53.74 & 59.94 & 60.72 \\

EEVG~\textbf{(Ours) } & DarkNet53&\Checkmark & 70.66 & 74.16 & 65.60 & 60.76 & 65.38 & 50.70 & 59.79 & 59.93 \\

EEVG~\textbf{(Ours) } & Swin-B&\Checkmark & 75.79 & 77.86 & 72.78 & 67.62 & 71.48 & 59.12 & 67.40 & 67.30 \\

EEVG~\textbf{(Ours) } & ViT-B&\Checkmark  & \textbf{78.23} & \textbf{79.27} & \textbf{76.58} & \textbf{69.04} & \textbf{72.65} & \textbf{62.33} & \textbf{69.15} & \textbf{70.01} \\

\noalign{\hrule height 1.5pt}
\end{tabular}
\label{tab:main_res_seg}
\end{table*}

\section{Experiments}
\subsection{Experiment Settings}

\textbf{Datasets. }
The commonly used datasets in visual grounding are   RefCOCO~\cite{refcoco_and_plus}, RefCOCO+~\cite{refcoco_and_plus}, and RefCOCOg~\cite{mao2016generation}, which are collected from MS-COCO~\cite{lin2014microsoft}. 
RefCOCO contains 19,994 images with 142,210 referring expressions for 50,000 objects which is split into the training set, the validation set, the testA set, and the testB set. 
RefCOCO+, excluding absolute-location words, consists of 19,992 images with 49,856 referred objects and 141,564 referring expressions.
There are 25,799 images with 49,856 referred objects and 141,564 referring expressions in RefCOCOg whose descriptions are longer and more complex. We use the umd-splits~\cite{nagaraja2016modeling} for RefCOCOg. Implementation details can be found in the Appendix.

\textbf{Evaluation Metrics. } For REC, we utilize the accuracy of the grounding results as the evaluation metric. The predicted region is deemed correct if the intersection over union (IoU) between the predicted region and the ground truth exceeds 0.5. As for RES, we employ the mean Intersection over Union (mIoU) between predicted masks and ground truth as the evaluation metric.

\subsection{Quantitative Results}

\begin{wraptable}[8]{r}{7cm}
\centering
\scriptsize
\caption{Speed comparison with SOTA methods. ``$\downarrow$'' means lower is better, ``$\uparrow$'' refers to upper is better, and ``FPS'' denotes frames per second. The batch size is 20 and all experiments are conducted in one RTX 4090. }
\begin{tabular}{c|ccc}
\noalign{\hrule height 1.5pt} 
Method & LAVT~\cite{yang2022lavt}  & PolyFormer~\cite{liu2023polyformer} & \textbf{Ours} \\
\hline

Runtime (ms) $\downarrow$ & 285.97 &  318.36 & \textbf{248.35} \\
\hline
FPS $\uparrow$ & 69.94  & 62.82 & \textbf{80.53} \\

\noalign{\hrule height 1.5pt}
\end{tabular}
\label{tab:speed_sota}
\end{wraptable}

    \textbf{Results of RefCOCO Series. } To validate the effectiveness of our method, we conduct experiments and report our performance on RefCOCO series datasets (\ie, RefCOCO/+/g). The results of REC and RES are reported in Table~\ref{tab:main_rec_det} and Table~\ref{tab:main_res_seg}, respectively. Our method outperforms previous state-of-the-art approaches in both REC and RES. 
    Particularly on RefCOCOg, which includes longer and more complex language expressions, our method exhibits even greater improvement (3.62\% in the val set and 3.93\% in the test set), showcasing its effectiveness in handling intricate scenes.

    \textbf{Speed Comparison. } In order to demonstrate the efficiency of our proposed method, as shown in Table~\ref{tab:speed_sota}, we conduct a speed comparison with LAVT~\cite{yang2022lavt} and PolyFormer~\cite{liu2023polyformer}. We use the same settings with LAVT and PolyFormer, \ie, Swin-B \& BERT-base as backbones, and the length of linguistic tokens is 20. To ensure a fair comparison, we exclude the time cost of point generation in PolyFormer, as it utilizes a point sequence approach to generate masks after visual-linguistic feature alignment. Compared with PolyFormer which utilizes Encoder for modalities fusion, we are faster than it $28.19\%$ in FPS (80.53 \textit{versus} 62.82). Compared with LAVT which devises an interaction module to fuse linguistic features in the visual backbone, we are still faster than it $15.14\%$ in FPS (80.53 \textit{versus} 69.94). 

    \textbf{Results of Pre-trained setting. }Table~\ref{tab:pretrain_sota} presents the results of our proposed method pre-trained on a large corpus of visual referring expression data and fine-tuned on the RefCOCO, RefCOCO+, and RefCOCOg datasets. The large corpus consists of the combination of Visual Genome~\cite{krishna2017visual}, RefCOCO~\cite{refcoco_and_plus}, RefCOCO+~\cite{refcoco_and_plus}, RefCOCOg~\cite{mao2016generation}, and Flickr30k~\cite{plummer2015flickr30k} datasets. Following PolyFormer~\cite{liu2023polyformer}, we pre-train our model using the REC task on this large corpus and subsequently fine-tune on the combined training sets of RefCOCO, RefCOCO+, and RefCOCOg in both REC and RES task, with all validation and test images removed. For the RES task, our method achieves state-of-the-art performance across all validation and test splits, outperforming previous methods by a considerable margin. In the REC task, our approach demonstrates superior or comparable performance to prior state-of-the-art techniques. 

    \begin{table*}[t]
\centering
\scriptsize

\caption{Comparison with pre-trained state-of-the-art methods on RefCOCO~\cite{refcoco_and_plus}, RefCOCO+~\cite{refcoco_and_plus}, and RefCOCOg~\cite{nagaraja2016modeling} for \textbf{REC} and \textbf{RES} task. \textbf{Bold} denotes the best performance. Swin-B and ViT-B are Swin-Transformer Base and ViT Base.}
\begin{tabular}{c|c|c|ccc|ccc|cc}
\noalign{\hrule height 1.5pt}
\multicolumn{1}{c|}{\multirow{2}{*}{Method}} & \multicolumn{1}{c|}{\multirow{2}{*}{Backbone}} & \multicolumn{1}{c|}{\multirow{2}{*}{Task Type}} & \multicolumn{3}{c|}{RefCOCO} & \multicolumn{3}{c|}{RefCOCO+} & \multicolumn{2}{c}{RefCOCOg} \\
\cline{4-11}
& & & val & test A & test B & val & test A & test B & val(U) & test(U) \\

\hline
RefTr~\cite{li2021referring} & ResNet101& REC & 85.65 & 88.73 & 81.16 & 77.55 & 
82.26 & 68.99 & 79.25 & 80.01  \\
SeqTR~\cite{zhu2022seqtr} &DarkNet53 & REC & 87.00& 90.15 & 83.59 & 78.69 & 84.51 & 71.87 & 82.69 & 83.37 \\
PolyFormer~\cite{liu2023polyformer} & Swin-B & REC & 89.73 & 91.73 & 86.03 & \textbf{83.73} & \textbf{88.60} & \textbf{76.38} & 84.46 & \textbf{84.96} \\
\hline
EEVG~\textbf{(Ours) }  & Swin-B & REC & 89.63 & 92.00 & 86.40 & 82.24 & 87.34 & 74.00 & 83.99 & 84.53 \\
EEVG~\textbf{(Ours) }  & ViT-B & REC & \textbf{90.47} & \textbf{92.73} & \textbf{87.72} & 81.79 & 87.80 & 74.94 & \textbf{85.19} & 84.72 \\

\noalign{\hrule height 1.5pt}
RefTr~\cite{li2021referring} & ResNet101 & RES & 74.34 & 76.77 & 70.87 & 66.75 & 70.58 & 59.40 & 66.63 & 67.39 \\
SeqTR~\cite{zhu2022seqtr} & DarkNet53 & RES & 71.70 & 73.31 & 69.82 & 63.04 & 66.73 & 58.97 & 64.69 & 65.74 \\
PolyFormer~\cite{liu2023polyformer} &Swin-B & RES & 75.96 & 77.09 & 73.22 & 70.65 & 74.51 & 64.64 & 69.36 & 69.88 \\
\hline
EEVG~\textbf{(Ours) }  & Swin-B & RES & 77.52 & 79.63 & 75.25 & 71.35 & 75.56 & 64.58 & 71.48 & 71.90 \\
EEVG~\textbf{(Ours) }  & ViT-B & RES & \textbf{79.49} & \textbf{80.87} & \textbf{77.39} & \textbf{71.86} & \textbf{76.67} & \textbf{66.31} & \textbf{73.56} & \textbf{73.47} \\
\noalign{\hrule height 1.5pt}
\end{tabular}
\label{tab:pretrain_sota}

\end{table*}

    \begin{table*}[t]
\centering
\scriptsize
\caption{Performance between Encoder and Decoder for cross-modal interaction.}

\begin{tabular}{c|c|c|ccc|ccc|cc}
\noalign{\hrule height 1.5pt}
Cross-modal  & \multicolumn{1}{c|}{\multirow{2}{*}{Backbone}}& \multicolumn{1}{c|}{\multirow{2}{*}{Task Type}} & \multicolumn{3}{c|}{RefCOCO} & \multicolumn{3}{c|}{RefCOCO+} & \multicolumn{2}{c}{RefCOCOg} \\
\cline{4-11}
Module & &  &val & test A & test B & val & test A & test B & val(U) & test(U) \\

\hline
Encoder & ViT-B& REC & 86.59 & 89.59 & 84.43 & 75.82 & 81.11 & 68.48 & 78.43 & 78.08 \\
Decoder & ViT-B& REC & \textbf{87.55} & \textbf{90.03} & \textbf{84.71} & \textbf{77.31} & \textbf{81.84} & \textbf{68.89} & \textbf{79.27} & \textbf{79.26} \\

\hline
Encoder & ViT-B & RES & 76.97 & 78.72 & 75.53 & 67.40 & 71.25 & 60.70 & 68.12 & 68.19 \\
Decoder & ViT-B & RES & \textbf{77.49} & \textbf{78.99} & \textbf{75.73} & \textbf{68.19} & \textbf{71.53} & \textbf{61.03} & \textbf{68.16} & \textbf{68.60} \\
\noalign{\hrule height 1.5pt}
\end{tabular}
\label{tab:encoder_decoder}
\end{table*}

\subsection{Ablation Studies}
    
    \begin{table*}[t]
\centering
\scriptsize

\caption{Runtime (ms) comparison between Encoder and Decoder, both of them with 3 layers. The batch size is 20 and all experiments are conducted in one RTX 4090. N and L denote the number of visual and linguistic tokens, respectively. The lower value is better in the table.}
\begin{tabular}{c|ccccc|ccccc}
\noalign{\hrule height 1.5pt}
\textit{N} (Visual Tokens) &   \multicolumn{5}{c|}{196} & \multicolumn{5}{c}{784} \\
\cline{2-11}
\textit{L} (Linguistic Tokens) & 60 & 100 & 150 & 200 & 300  & 60 & 100 & 150 & 200 & 300 \\
\hline
Encoder  & 6.41 & 7.99 & 9.64 & 11.14 & 14.85  & 28.56 & 30.33 & 33.12 & 35.39 & 40.51 \\
Decoder  & \textbf{5.80} & \textbf{6.06} & \textbf{6.59} & \textbf{6.93} & \textbf{8.09}  & \textbf{28.10} & \textbf{28.74} & \textbf{29.93} & \textbf{30.81} & \textbf{33.00} \\

\noalign{\hrule height 1.5pt}
\end{tabular}
\label{tab:speed_encoder_decoder}
\end{table*}

    \textbf{Comparison between Encoder and Decoder. } To demonstrate the effectiveness and efficiency of Decoder, we conduct experiments to compare Decoder
    \begin{wraptable}[13]{r}{6.75cm}
\centering
\scriptsize

\caption{Ablation study of eliminating tokens on RefCOCOg. In the static eliminating strategy, we remove 96 tokens from each layer. ASA denotes adaptive spatial attention, ``$\downarrow$'' means lower is better, and ``$\uparrow$'' refers to higher is better. The runtime of token elimination module is tested in one RTX 4090 and the batch size is 20.}
\begin{tabular}{c|c|c|cc|cc}
\noalign{\hrule height 1.5pt} 
Elimination  & \multirow{2}{*}{ASA} & Runtime  & \multicolumn{2}{c|}{RES} & \multicolumn{2}{c}{REC}\\
\cline{4-7}
 Strategy&  & (ms) $\downarrow$ &val $\uparrow$ & test $\uparrow$ & val $\uparrow$ & test $\uparrow$ \\
\hline


 No & \XSolidBrush & 27.85 & 68.16 & 68.60 & 79.27 & 79.26  \\
  Static & \XSolidBrush & 24.72 & 66.38 & 66.94 & 78.33 &  79.18 \\
  Static & \Checkmark & 24.88 & 67.75 & 68.21 & 78.72 & 79.20  \\
Dynamic & \XSolidBrush & \textbf{24.68} & 68.89 & 68.95 & 79.27 & 79.34  \\
  Dynamic & \Checkmark & 24.83 & \textbf{69.15} & \textbf{70.01} & \textbf{79.60} & \textbf{80.24}  \\

\noalign{\hrule height 1.5pt}
\end{tabular}
\label{tab:eliminating}
\end{wraptable}
    with Encoder for vision language fusion. Table~\ref{tab:encoder_decoder} demonstrates that Decoder outperforms Encoder, highlighting Decoder's ability to facilitate vision language alignment. Additionally, Table~\ref{tab:speed_encoder_decoder} indicates that Decoder exhibits faster speed compared to Encoder, particularly when handling longer language expressions. It should be noted that Decoder incorporates an additional multi-head cross-attention module. To ensure a fair comparison, we set the dimension of the feed-forward network to 1024 in Decoder and 2048 in Encoder, thereby aligning their parameter quantities.

\begin{wraptable}[11]{r}{7cm}
\scriptsize
\caption{Ablation study of mask head on RefCOCOg. The FPN-like is the mask head based on convolution layers within the FPN framework, which has prevailed in recent VG works~\cite{feng2021encoder,yang2022lavt,li2021referring,su2023vglaw}. ``$\downarrow$'' means lower is better and ``$\uparrow$'' refers to higher is better. The runtime of mask head is tested in one RTX 4090 and batch size is 20. }
\begin{tabular}{c|c|c|cc|cc}
\noalign{\hrule height 1.5pt} Mask & Parameter &Runtime & \multicolumn{2}{c|}{RES} & \multicolumn{2}{c}{REC}\\
\cline{4-7}
Head & Number $\downarrow$ & (ms) $\downarrow$ & val $\uparrow$ & test $\uparrow$ & val $\uparrow$ & test $\uparrow$ \\
\hline

FPN-like & 8.11M & 23.56 & 68.64 & 69.12 & 77.68 & 78.26  \\
\textbf{Ours} & \textbf{0.79M} & \textbf{0.87} & \textbf{69.15} & \textbf{70.01} & \textbf{79.60} &  \textbf{80.24} \\

\noalign{\hrule height 1.5pt}
\end{tabular}
\label{tab:mask_head}
\end{wraptable}

    \textbf{Token Elimination. } To validate the advantages of our token elimination strategy, we compare it with the results obtained without elimination as well as the results achieved using a static elimination strategy~\cite{fayyaz2022adaptive}. 
    As depicted in Table~\ref{tab:eliminating}, employing the static token elimination strategy leads to a decline in performance when compared to the absence of an elimination strategy. Conversely, the dynamic token elimination strategy not only reduces computational costs but also exhibits performance improvements. Furthermore, the adoption of adaptive spatial attention demonstrates enhancement in performance, thereby mitigating the issue of erroneously eliminating certain visual tokens associated with target objects, as illustrated in Fig.~\ref{fig:attention_score}.

    \textbf{Efficient Mask Head. } In comparison to the FPN-like mask head~\cite{kirillov2019panoptic}, as reported in Table~\ref{tab:mask_head}, our mask head has fewer parameters and faster speed. Moreover, our mask head not only demonstrates improvements in the RES task but also exhibits enhancements in the REC task. We believe this is because our light mask head offloads the spatial prediction workload from the task head add-ons to Decoder which is consistent with the light MLP-based detection head. Consequently, Decoder benefits from improved vision-language fusion, as the location and pixel information are now embedded within the Decoder token feature channels.

\subsection{Qualitative Results}
    \begin{figure}[t]
        \centering
        \includegraphics[width=1\linewidth]{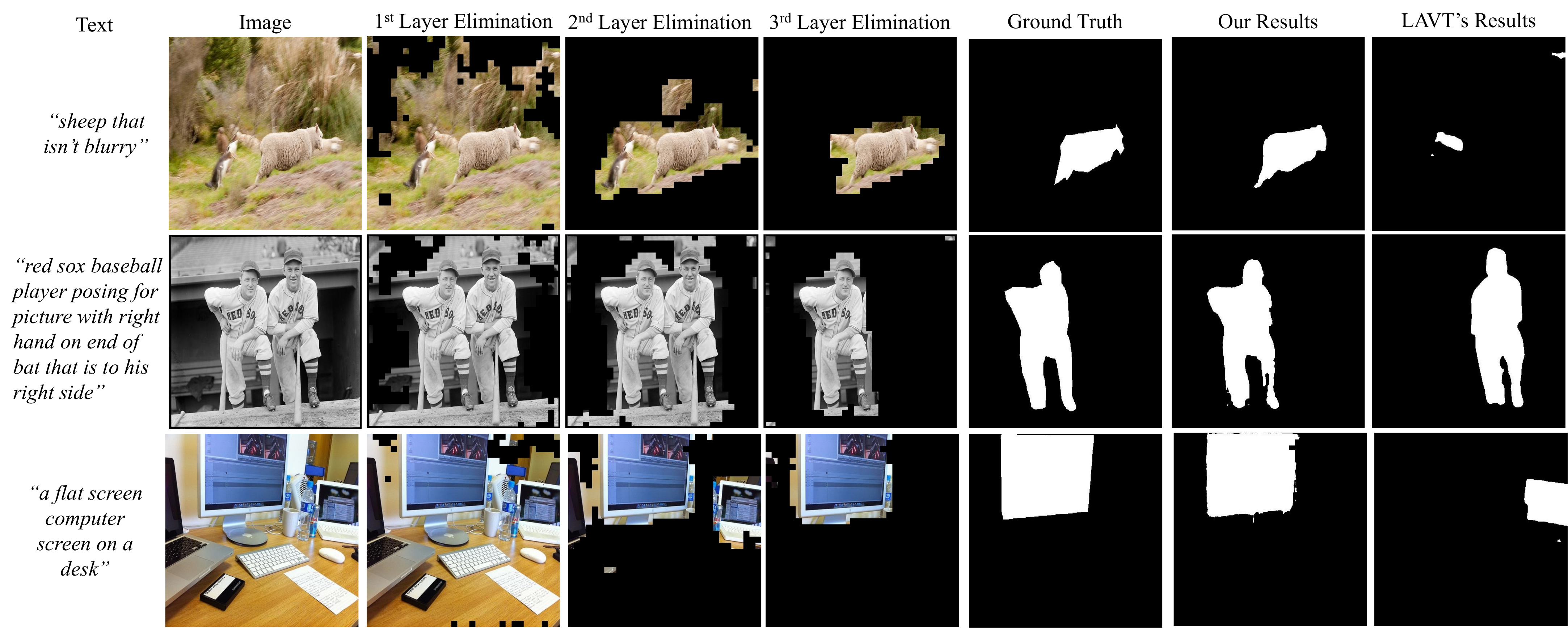}
        \caption{Visualization of our eliminating process, our predicted results, and LAVT's~\cite{yang2022lavt}. }
        \label{fig:qualitive_analysis}
    \end{figure}
    
    We present qualitative results achieved using our proposed method and compare them with the results obtained from LAVT, as shown in Fig.~\ref{fig:qualitive_analysis}. These results demonstrate the effectiveness of our approach in managing complex scenes with similar and potentially distracting objects. In such situations, LAVT tends to make incorrect predictions because of these distracting objects. However, our method removes these distracting objects in various Decoder layers, allowing for accurate predictions of the target object. More visualization examples can be found in the Appendix.

\section{Conclusion}
In this paper, we present a novel approach to visual grounding that achieves superior performance while requiring less computational resources. By incorporating visual and linguistic features through cross-attention in the Transformer Decoder, our method effectively handles longer language expressions without significantly increasing the computational cost. To further enhance efficiency, we introduce a parameter-free strategy to remove unnecessary visual tokens during cross-modal fusion. This strategy not only reduces computations but also improves overall performance by eliminating irrelevant objects. After obtaining sparse visual tokens, we propose an efficient mask head that directly generates masks without the need for padding. Extensive experiments conducted on benchmark datasets validate that our method surpasses state-of-the-art techniques in both referring expression comprehension and segmentation tasks.

\section*{Acknowledgment}
This work was partially supported by the National Natural Science Foundation of China under Grant 62372341.
Long Chen was supported by HKUST Special Support for Young Faculty (F0927) and HKUST Sports Science and Technology Research Grant (SSTRG24EG04).

%
%
\bibliographystyle{splncs04}
\bibliography{main}

\newpage
\section*{Appendix}

\begin{wrapfigure}{r}{6cm}
    \centering
    \includegraphics[width=1\linewidth]{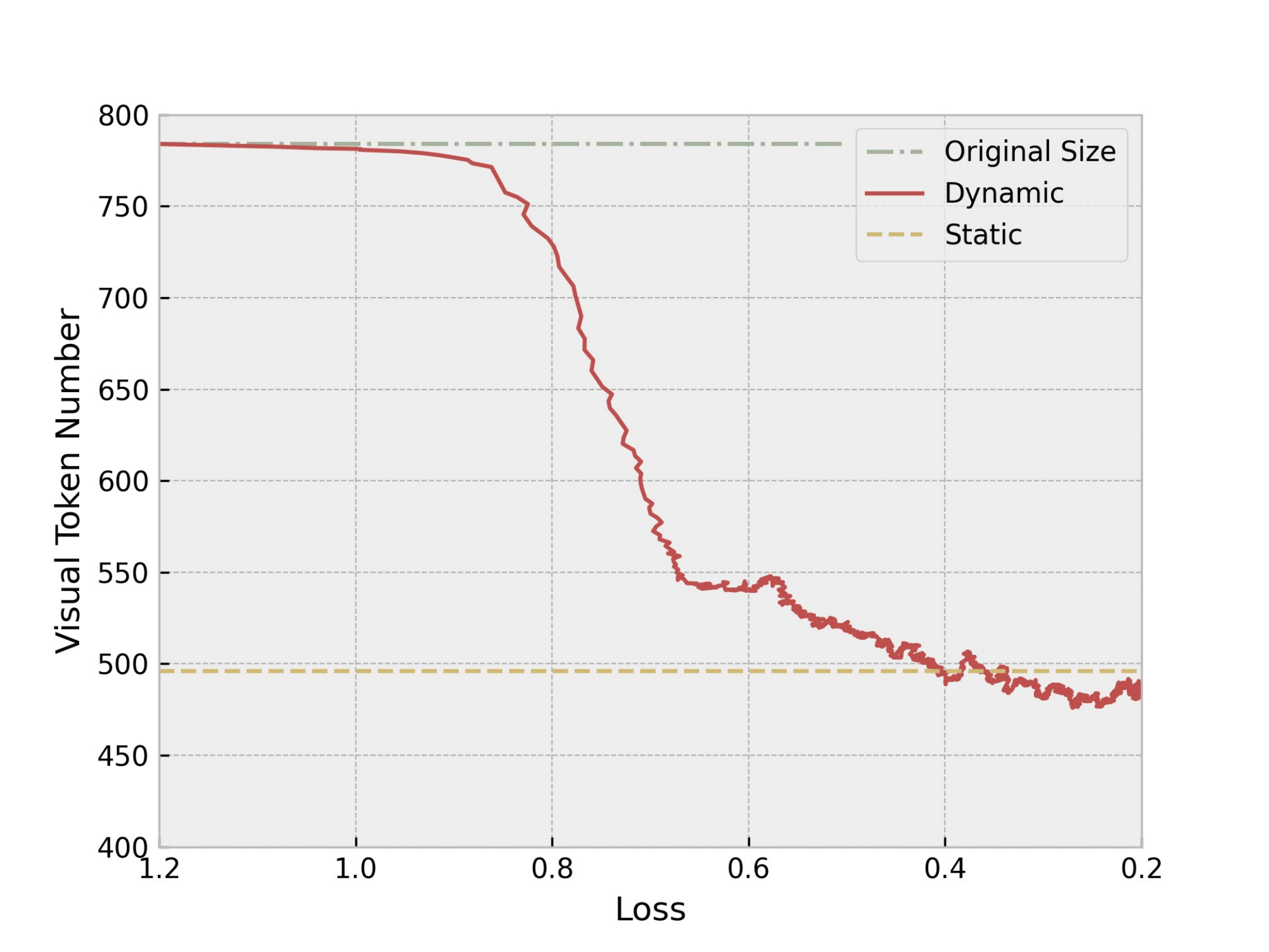}
    \caption{Visualization of eliminating visual tokens process. }
    \label{fig:loss_token_num}
\end{wrapfigure}

\section{More Analysis of Elimination Strategy}
Compared to the static elimination strategy, our dynamic elimination strategy offers an improved approach to address the issue of incorrect elimination during the training process. The static strategy eliminates a fixed number of visual tokens, which can lead to premature elimination and negatively impact the model's performance. In contrast, our dynamic strategy gradually increases the number of eliminated tokens as the loss converges, as shown in Figure~\ref{fig:loss_token_num}. This prevents incorrect elimination and ensures a more stable training process. This is why our strategy outperforms the static elimination strategy in the results of previous ablation studies.

\section{More Analysis of Mask Head}
\begin{wraptable}[8]{r}{6cm}
\centering
\caption{Ablation study results of convolution layer in our mask head. ``w/o Conv'' denotes without using convolution layer. }
\begin{tabular}{c|cc|cc}
\noalign{\hrule height 1.5pt} \multirow{2}{*}{Type} & \multicolumn{2}{c|}{RES} & \multicolumn{2}{c}{REC}\\
\cline{2-5}
  & val & test  & val & test \\
\hline

w/o Conv & 67.71 & 68.18  & 78.95 & 79.00 \\
with Conv & \textbf{69.15} & \textbf{70.01} & \textbf{79.60} & \textbf{80.24} \\

\noalign{\hrule height 1.5pt}
\end{tabular}
\label{tab:ablation_conv}
\end{wraptable}

\begin{figure*}
    \centering
    \includegraphics[width=1\linewidth]{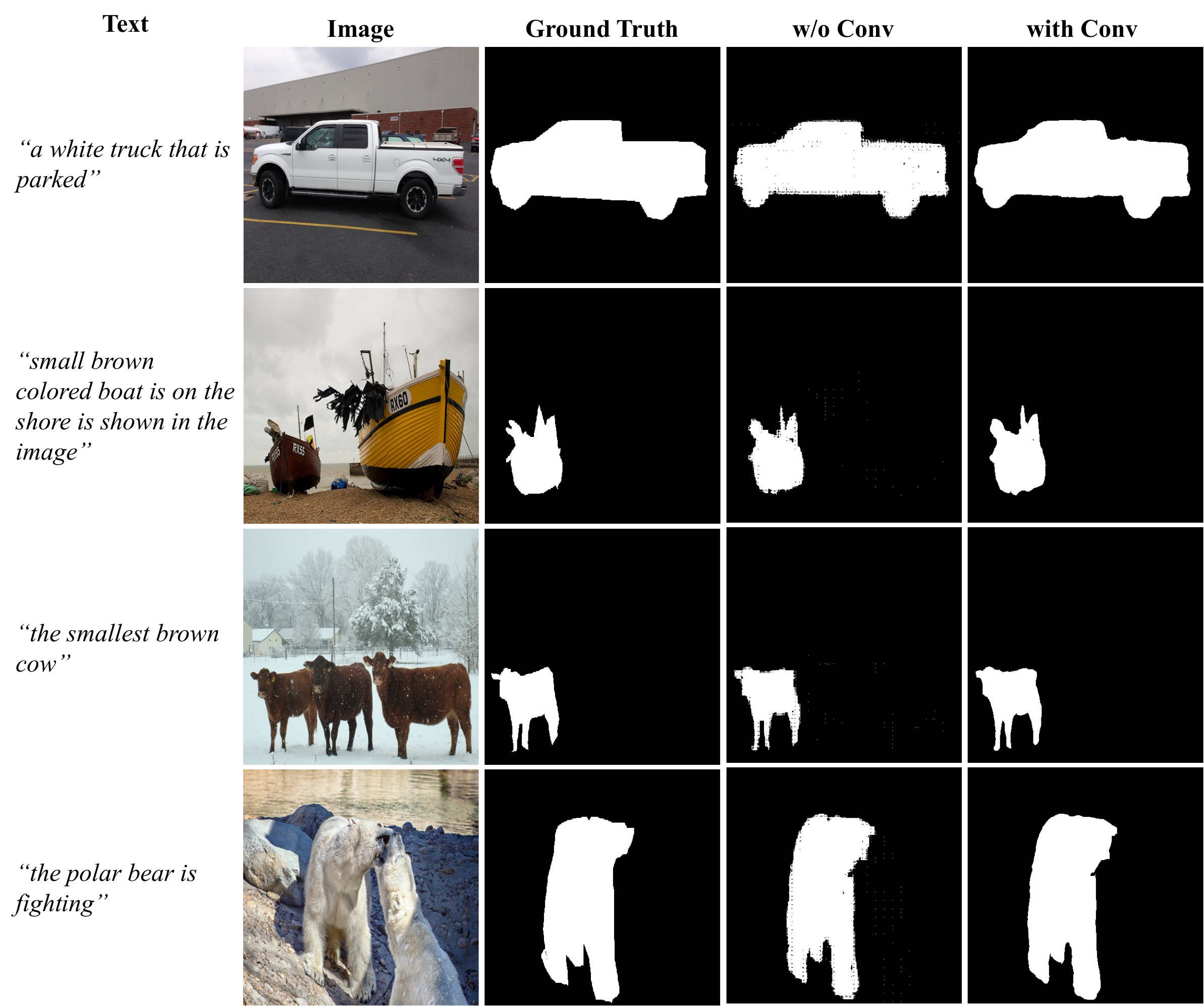}
    \caption{Comparison of our predicted results between  ``w/o Conv'' and ``with Conv''. ``w/o Conv'' means without using convolution layer in our mask head. }
    \label{fig:fig_conv}
\end{figure*}

We observed that when using an MLP to project visual tokens into a spatial segmentation mask, the resulting mask lacks spatial relationships between pixels, as depicted in Figure~\ref{fig:fig_conv}. In order to address this issue, we propose the incorporation of a 1-channel convolution layer. This layer helps establish connections between neighboring pixels, thus enhancing the spatial coherence of the resulting mask. The ablation study results are shown in Table~\ref{tab:ablation_conv}.



\section{Additional Qualitative Results}
\begin{figure*}
    \centering
    \includegraphics[width=1\linewidth]{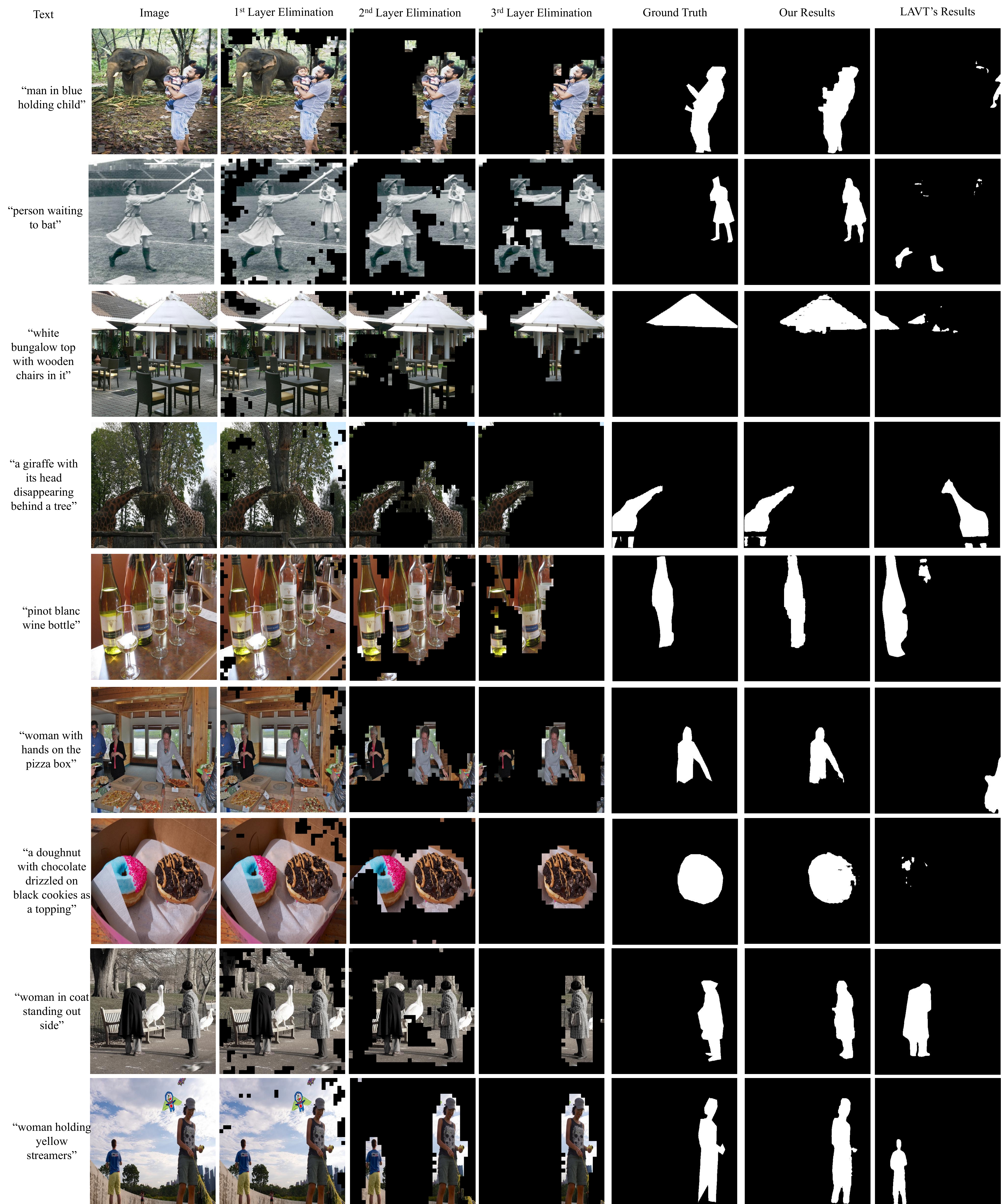}
    \caption{Additional visualization of our results. }
    \label{fig:addition_results}
\end{figure*}

As shown in Figure~\ref{fig:addition_results}, we provide further visualization examples that illustrate the process of our method for eliminating visual tokens as well as compare the predicted results of our approach with those of LAVT~\cite{yang2022lavt}.

\section{Generalization on GRES.}
\begin{wraptable}[8]{r}{8cm}
    \tabcolsep=0.5mm
    \centering
    \caption{Performance comparison on GRES~\cite{liu2023gres}}
    \begin{tabular}{c|cc|cc|cc}
    \noalign{\hrule height 1.5pt}
    \multirow{2}{*}{Methods} & \multicolumn{2}{c|}{val} & \multicolumn{2}{c|}{testA} & \multicolumn{2}{c}{testB} \\
    \cline{2-7}
     &cIoU & gIoU & cIoU & gIoU & cIoU & gIoU\\
    \hline
    LAVT [\textcolor{green}{42}] & 57.64 & 58.40 & 65.32 & 65.90 & 55.04  & 55.83\\
    ReLA~\cite{liu2023gres} & 62.42 & \textbf{63.60}  & 69.26 & 70.03 & 59.88 & 61.02\\
    \textbf{Ours} & \textbf{64.04} & 62.75 & \textbf{71.65} & \textbf{70.93} & \textbf{62.77} & \textbf{62.79}\\
    \noalign{\hrule height 1.5pt}
    \end{tabular}
    \label{tab:rec_res}
\end{wraptable}
We further validate our approach using the GRES~\cite{liu2023gres} dataset, which includes text corresponding to either multiple objects or none. In this new scenario, we also achieve SOTA, demonstrating the superiority and robustness of our method.

\section{Implementation Details. }
We use the AdamW optimizer~\cite{loshchilov2018decoupled} and the weight decay is 1e-4. The initial learning rate is 5e-6 for the language backbone, 1e-5 for the visual backbone, and 2.5e-5 for the rest of the model. BERT-base~\cite{devlin2018bert} is utilized as the linguistic backbone for extracting linguistic features, while ViT-base~\cite{dosovitskiy2020image} serves as the visual backbone. We employ the adaptation introduced by ViTDet~\cite{li2022vitdet} to adjust the visual backbone for higher-resolution images (\ie, $448 \times 448$ in our method), and it is pre-trained on MS-COCO~\cite{lin2014microsoft}, excluding overlapping images from the val/test sets. The number of Transformer Decoder layers is 3, the hidden dimension is 768, and the feed-forward network dimension is 1024. We train the model for 150 epochs with a batch size of 80 using RTX 4090s. The patch size $P$ is 16, 
the threshold $\alpha$ is 0.015, $k$ in adaptive spatial attention is 1, the convolutional kernel is 5*5 in our mask head, and $\lambda_{det}$ and $\lambda_{seg}$ are 0.1 and 1. 

\end{document}